\documentclass[twocolumn]{article}

\usepackage{amsmath,amssymb}
\usepackage{algorithmic}
\usepackage{algorithm}
\usepackage{theorem}
\usepackage{amsmath}
\usepackage{amssymb}

\usepackage[pagebackref=true,breaklinks=true,letterpaper=true,colorlinks,bookmarks=false]{hyperref}

\newcommand{\nbr}[1]{\left\|#1\right\|}
\DeclareMathOperator*{\argmax}{\mathrm{argmax}}
\DeclareMathOperator*{\argmin}{\mathrm{argmin}}

\usepackage{cite}
\usepackage{graphicx}
\usepackage{url}

\begin{document}
\title{Robust Near-Isometric Matching via Structured Learning of Graphical Models}

\author{Julian~J.~McAuley\thanks{The authors are with the Statistical Machine Learning Program at NICTA, and the Research School of Information Sciences and Engineering, Australian National University.}, Tib\'erio~S.~Caetano and Alexander J. Smola}

\maketitle

\begin{abstract}

Models for near-rigid shape matching are typically based on distance-related features, in order to infer matches that are consistent with the isometric assumption. However, real shapes from image datasets, even when expected to be related by ``almost isometric'' transformations, are actually subject not only to noise but also, to some limited degree, to variations in appearance and scale. In this paper, we introduce a graphical model that parameterises appearance, distance, and angle features and we learn all of the involved parameters via structured prediction. The outcome is a model for near-rigid shape matching which is \emph{robust} in the sense that it is able to capture the possibly limited but still important scale and appearance variations. Our experimental results reveal substantial improvements upon recent successful models, while maintaining similar running times.

\end{abstract}

\section{Introduction}

Matching shapes in images has many applications, including image retrieval, alignment, and registration \cite{BelMalPuz02,MorBelMal01,mori02estimating,frome04recognizing}. Typically, matching is approached by selecting features for a set of landmark points in both images; a correspondence between the two is then chosen such that some distance measure between these features is minimised. A great deal of attention has been devoted to defining complex features which are robust to changes in rotation, scale etc. \cite{BelMal00,lowe99object}.\footnote{We restrict our attention to this type of approach, i.e.~that of matching landmarks between images. Some notable approaches deviate from this norm -- see (for example) \cite{fel05,fel07,LeHuaBot04,CarHeb04}.}

An important class of matching problems is that of \emph{near-isometric} shape matching. In this setting, it is assumed that shapes are defined up to an isometric transformation (allowing for some noise), and therefore distance features are typically used to encode the shape. Some traditional methods for related settings focus on optimisation over the space of rigid transformations so as to minimise least-squares criteria \cite{HarZis04,Fitzgibbon01}. 

Recently, this class of problems has been approached from a different perspective, as direct optimisation over the space of correspondences \cite{CaeCaeSchbar06}. Although apparently more expensive, there it is shown that the rigidity assumption imposes a convenient algebraic structure in the correspondence space so as to allow for efficient algorithms (exact inference in chordal graphical models of small clique size). More recently, these methods have been made substantially faster \cite{McACaeBar08}. The \emph{key idea} in these methods is to \emph{explicitly encode rigidity constraints} into a tractable graphical model whose MAP solution corresponds to the best match. However, the main advantages of correspondence-based optimisation over transformation-based optimisation, namely the flexibility of encoding powerful local features, has not been further explored in this framework.

Other lines of work that optimise directly over the correspondence space are those based on Graph Matching, which explicitly model all pairwise compatibilities and solve for the best match with some \emph{relaxation} (since the Graph Matching problem is NP-hard for general pairwise compatibilities) \cite{CouSriShi06,berg-etal05,Leor05}. Recently, it was shown both in \cite{CaeCheLeSmo07} and in \cite{LeoHeb08} that if some form of \emph{structured optimisation} is used to optimise graph matching scores, relaxed quadratic assignment predictors can improve the power of pairwise features. The \emph{key idea} in these methods is to \emph{learn the compatibility scores} for the graph matching objective function, therefore enriching the representability of features. A downside of these graph matching methods however is that they do not typically make explicit use of the geometry of the scene in order to improve computational efficiency and/or accuracy.

In this paper, we combine these two lines of work into a single framework. We produce an exact, efficient model to solve near-isometric shape matching problems using not only isometry-invariant features, but also appearance and scale-invariant features, all encoded in a \emph{tractable graphical model}. By doing so we can \emph{learn via large-margin structured prediction} the relative importances of variations in appearance and scale with regard to variations in shape \emph{per se}. Therefore, even knowing that we are in a near-isometric setting, we will still capture the eventual variations in appearance and scale into our matching criterion in order to produce a \emph{robust} near-isometric matcher. In terms of learning, we introduce a two-stage structured learning approach to address the speed and memory efficiency of this model.

The remainder of this paper is structured as follows: in section \ref{sec:background}, we give a brief introduction to shape matching (\ref{sec:shapematch}), graphical models (\ref{sec:graphmodels}), and discriminative structured learning (\ref{sec:learning}). In section \ref{sec:ourmodel}, we present our model, and experiments follow in section \ref{sec:experiments}.

\section{Background}
\label{sec:background}

\subsection{Shape Matching}
\label{sec:shapematch}

`Shape matching' can mean many different things, depending on the precise type of query one is interested in. Here we study the case of identifying an instance of a template shape ($\mathcal S \subseteq \mathcal T$) in a target scene ($\mathcal U$) \cite{BelMalPuz02}.\footnote{Here $\mathcal T$ is the set of \emph{all points} in the template scene, whereas $\mathcal S$ corresponds to those points in which we are interested. It is also important to note that we treat $\mathcal S$ as an \emph{ordered} object in our setting.} We assume that we \emph{know} $\mathcal S$, i.e.~the points in the template that we want to query in the scene. Typically both $\mathcal T$ and $\mathcal U$ correspond to a set of `landmark' points, taken from a pair of images (common approaches include \cite{lowe99object,Canny87,Deriche87,susan92}).

For each point $t \in \mathcal T$ and $u \in \mathcal U$, a certain set of \emph{unary features} are extracted (here denoted by $\phi(t)$, $\phi(u)$), which contain local information about the image at that point \cite{BelMal00,lowe99object}. If $y: \mathcal S \rightarrow \mathcal U$ is a generic mapping representing a potential match, the goal is then to find a mapping  $\hat{y}$ which minimises the aggregate distance between corresponding features, i.e.
\begin{equation}
 \hat{y} = f(\mathcal S,\mathcal U) = \argmin_y \sum_{i=1}^{|\mathcal S|} c_1(s_i, y(s_i))\label{eq:min1}
\end{equation}
where
\begin{equation}
 c_1(s_i, y(s_i)) = {\nbr{\phi(s_i) - \phi(y(s_i))}_2^2}.
\end{equation}
(here $\nbr{\cdot}_2$ denotes the $L_2$ norm). For injective $y$ eq.~(\ref{eq:min1}) is a linear assignment problem, efficiently solvable in cubic time.

In addition to unary or first-order features, pairwise or second-order features can be induced from the locations of the unary features. In this case eq.~(\ref{eq:min1}) is generalised to minimise an aggregate distance between pairwise features, i.e.
\begin{equation}
 \hat{y} = \argmin_y \sum_{i=1}^{|\mathcal S|} c_1(s_i, y(s_i)) + \sum_{i=1}^{|\mathcal S|}\sum_{j=1}^{|\mathcal S|} c_2(s_i, s_j, y(s_i), y(s_j)).
\label{eq:min2}
\end{equation}
This however induces an NP-hard problem for general $c_2$ (quadratic assignment). Discriminative structured learning has recently been applied to models of both linear and quadratic assignment (eq.~(\ref{eq:min1}) and eq.~(\ref{eq:min2})) in \cite{CaeCheLeSmo07}. Here we exploit the structure of $c_2$ that arises from the near-isometric shape matching problem in order to make such a problem tractable.

\subsection{Graphical Models}
\label{sec:graphmodels}

\begin{figure}
\centering
\includegraphics[height=5cm]{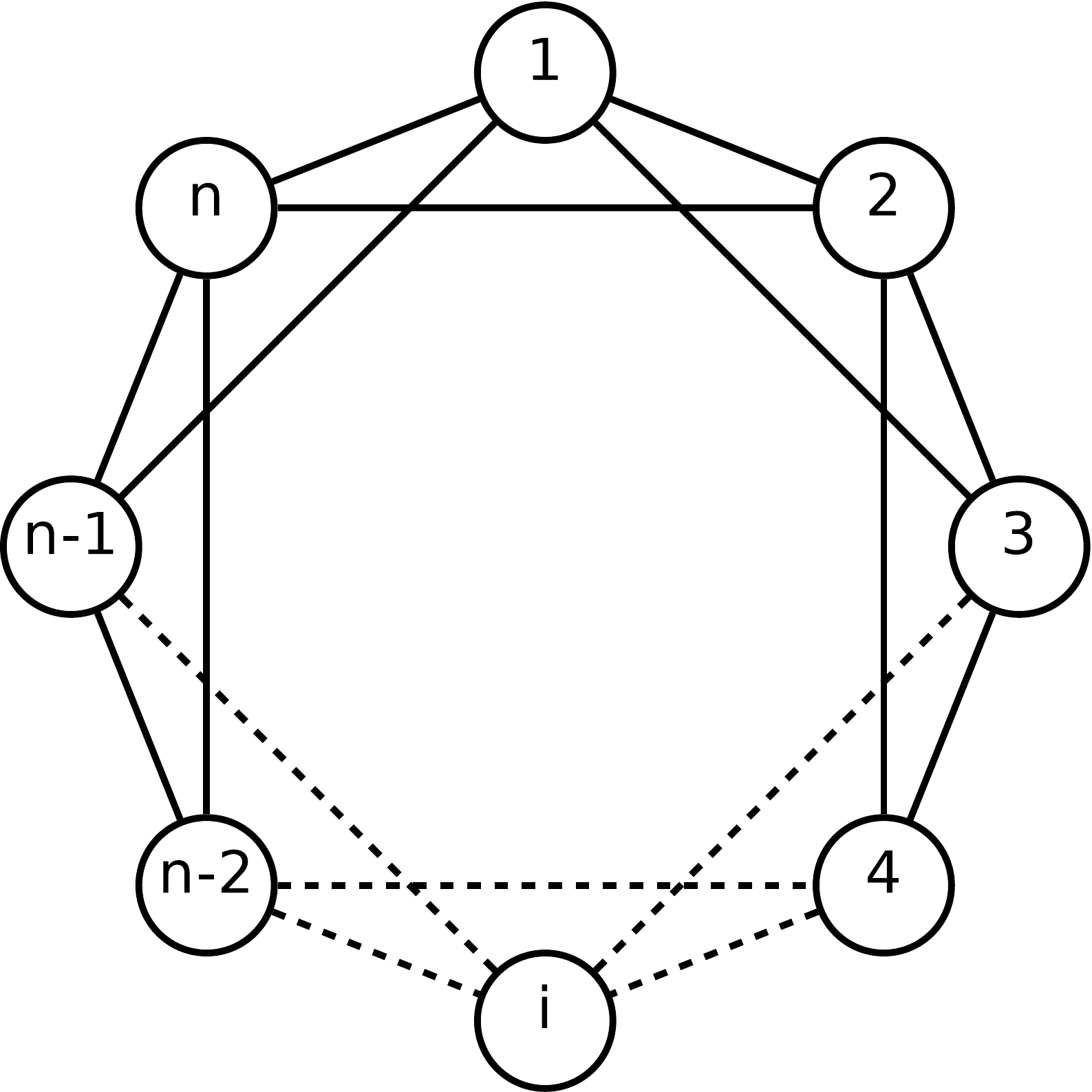}

\vspace{5mm}

\includegraphics[height=5cm]{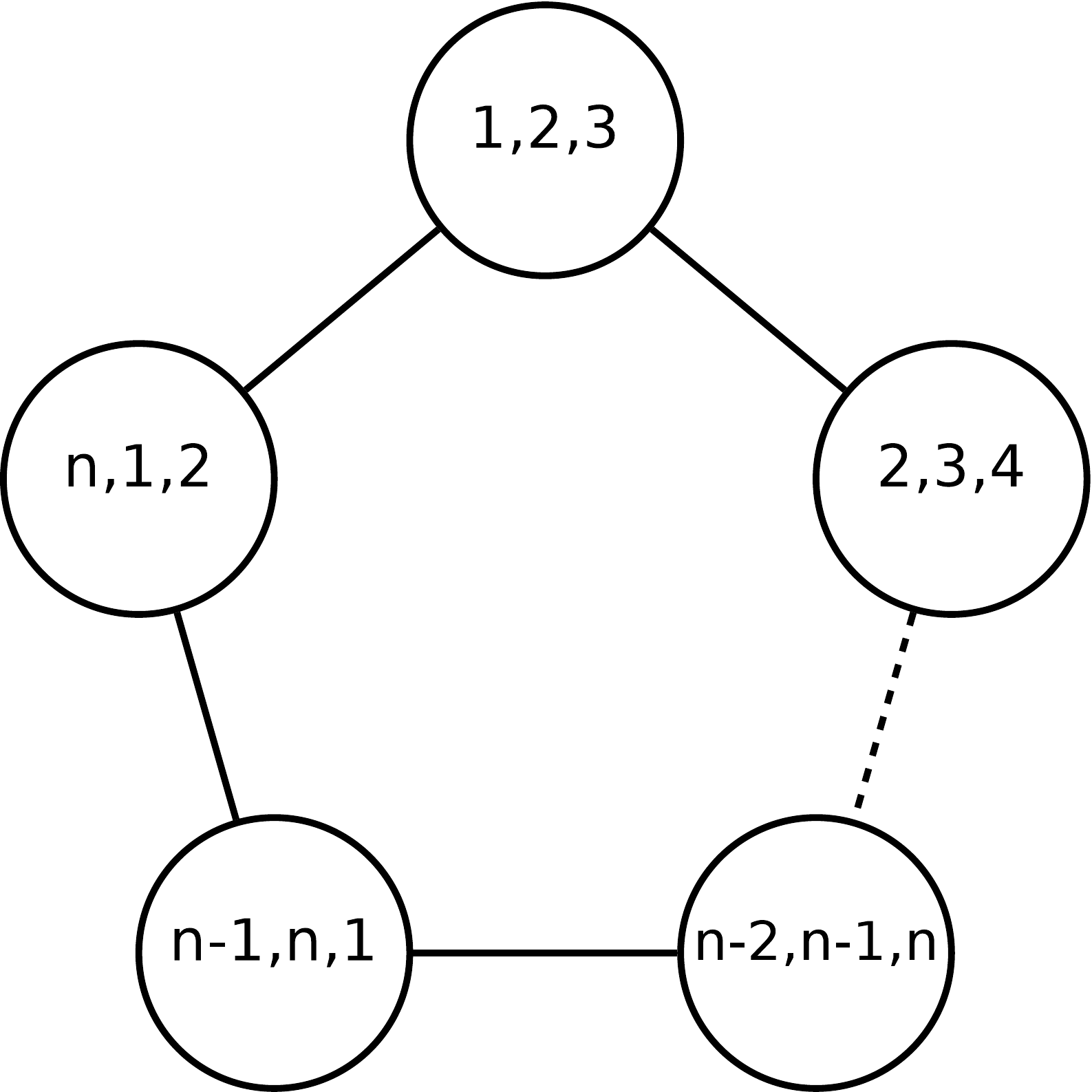}
\caption{Top: The graphical model introduced in \cite{McACaeBar08}. Bottom: The clique-graph of this model.}
\label{fig:model}
\end{figure}

In isometric matching settings, one may suspect that it may not be necessary to include all pairwise relations in quadratic assignment. In fact a recent paper \cite{McACaeBar08} has shown that if only the distances as encoded by the graphical model depicted in figure \ref{fig:model} (top) are taken into account (nodes represent points in $\mathcal S$ and states represent points in $\mathcal U$), exact probabilistic inference in such a model can solve the isometric problem optimally. That is, an energy function of the following form is minimised:\footnote{$s_{i+1}$ should be interpreted as $s_{(i+1)\mod |\mathcal S|}$ (i.e. the points form a loop).}
\begin{equation}
\sum_{i=1}^{|\mathcal S|} c_2(s_i, s_{i+1}, y(s_i), y(s_{i+1})) + c_2(s_i, s_{i+2}, y(s_i), y(s_{i+2})).
\label{eq:min3}
\end{equation}

Although the graphical model in figure \ref{fig:model} (top) does not form a single loop (a condition typically required for convergence of belief propagation \cite{IhlFisWil05,Weiss00,WeiFre01}), \cite{McACaeBar08} show that it is sufficient that the \emph{clique graph} forms a single loop in order to guarantee convergence to the optimal assignment (figure \ref{fig:model}, bottom). Furthermore, it is shown in \cite{McACaeBar08} that the number of iterations required before convergence is small in practice.

We will extend this model by including a unary term, $c_1(s_i, y(s_i))$ (as in (eq.~\ref{eq:min1})), as well as a third-order term, $c_3(s_i, s_{i+1}, s_{i+2}, y(s_i), y(s_{i+1}), y(s_{i+2}))$; the graph topology remains the same. Note that in order to guarantee convergence, we do not require any specific form for the potentials, except that no assignment has infinite cost \cite{McACaeBar08}.

\subsection{Discriminative Structured Learning}
\label{sec:learning}

In practice, feature vectors may be very high-dimensional, and which components are `important' will depend on the specific properties of the shapes being matched. Therefore, we introduce a parameter, $\theta$, which controls the relative importances of the various feature components. Note that $\theta$ is parameterising the matching criterion itself. Hence our optimisation problem becomes
\begin{equation}
 \hat{y} = f(\mathcal S, \mathcal U; \theta) = \argmax_y \langle h(\mathcal S, \mathcal U, y), \theta \rangle \label{eq:mintheta}
\end{equation}
where
\begin{equation}
h(\mathcal S, \mathcal U, y) =
-\sum_{i=1}^{|\mathcal S|} \Phi(s_i, s_{i+1}, s_{i+2},y(s_i), y(s_{i+1}), y(s_{i+2})).\label{eq:negcost}
\end{equation}
($y$ is a mapping from $\mathcal S$ to $\mathcal U$, $\Phi$ is a third-order feature vector -- our specific choice is shown in section \ref{sec:ourmodel}).\footnote{We have expressed (eq.~\ref{eq:mintheta}) as a maximisation problem as a matter of convention; this is achieved simply by negating the cost function in (eq.~\ref{eq:negcost}).} In order to measure the performance of a particular weight vector, we use a \emph{loss function}, $\Delta(\hat{y}, y^i)$, which represents the cost incurred by choosing the assignment $\hat{y}$ when the correct assignment is $y^i$ (our specific choice of loss function is described in section \ref{sec:experiments}). To avoid overfitting, we also desire that $\theta$ is sufficiently `smooth'. Typically, one uses the squared $L_2$ norm, $\nbr{\theta}_2^2$, to penalise non-smooth choices of $\theta$ \cite{Tsoch04}.

Learning in this setting now becomes a matter of choosing $\theta$ such that the empirical risk (average loss on all training instances) is minimised, but which is also sufficiently `smooth' (to prevent overfitting). Specifically, if we have a set of training pairs, $\left\lbrace (\mathcal S^1,\mathcal U^1), \ldots, (\mathcal S^N,\mathcal U^N)\right\rbrace$, with labelled matches $\left\lbrace y^1 \ldots y^N\right\rbrace$, then we wish to minimise
\begin{equation}
\label{eq:minloss}
 \underbrace{\frac{1}{N} \sum_{i=1}^N \Delta(f(\mathcal S^i, \mathcal U^i; \theta), y^i)}_{\mbox{empirical risk}} + \underbrace{\frac{\lambda}{2} \nbr{\theta}_2^2}_{\mbox{regulariser}}.
\end{equation}
Here $\lambda$ (the regularisation constant) controls the relative importance of minimising the empirical risk against the regulariser. In our case, we simply choose $\lambda$ such that the empirical risk on our validation set is minimised.

Solving (eq.~\ref{eq:minloss}) exactly is an extremely difficult problem and in practice is not feasible, since the loss is piecewise constant on the parameter $\theta$. Here we capitalise on recent advances in large-margin structured estimation \cite{Tsoch04}, which consist of obtaining convex relaxations of this problem. Without going into the details of the solution (see, for example, \cite{Tsoch04,TeoLeSmoVis07}), it can be shown that a convex relaxation of this problem can be obtained, which is given by
\begin{subequations}
  \label{eq:convex}
  \begin{align}
    \label{eq:primal-obj}
    & \min_{\theta} ~
     \frac{1}{N}\sum_{i=1}^N \xi_i + \frac{\lambda}{2} \nbr{\theta}^2_2 \\
    \nonumber
    & \text{subject to } \\
\nonumber
    &\langle h(\mathcal S^i, \mathcal U^i, y^i) - h(\mathcal S^i, \mathcal U^i, y), \theta \rangle \geq \Delta(y,y^i) - \xi_i \\
    \label{eq:convex-cons}
    & \text{for all }
    i\text{ and } y \in \mathcal Y
  \end{align}
\end{subequations}
(where $\mathcal Y$ is the space of all possible mappings). It can be shown that for the solution of the above problem, we have that $\xi^*_i \ge \Delta(f(\mathcal S^i,\mathcal U^i;\theta),y^i)$. This means that we end up minimising an upper bound on the loss, instead of the loss itself.

Solving (\ref{eq:convex}) requires only that we are able, for any value of $\theta$, to find
\begin{equation}
 \argmax_y \left( \langle h(\mathcal S^i, \mathcal U^i, y), \theta \rangle + \Delta(y, y^i) \right).
\label{eq:plusloss}
\end{equation}
In other words, for each value of $\theta$, we are able to identify the mapping which is consistent with the model (eq.~\ref{eq:mintheta}), yet incurs a high loss. This process is known as `column generation' \cite{Tsoch04,TeoLeSmoVis07}. As we will define our loss as a sum over the nodes, solving (eq.~\ref{eq:plusloss}) is no more difficult than solving (eq.~\ref{eq:mintheta}).

\section{Our Model}
\label{sec:ourmodel}

\begin{figure*}
\centering
\includegraphics[height=3.5cm]{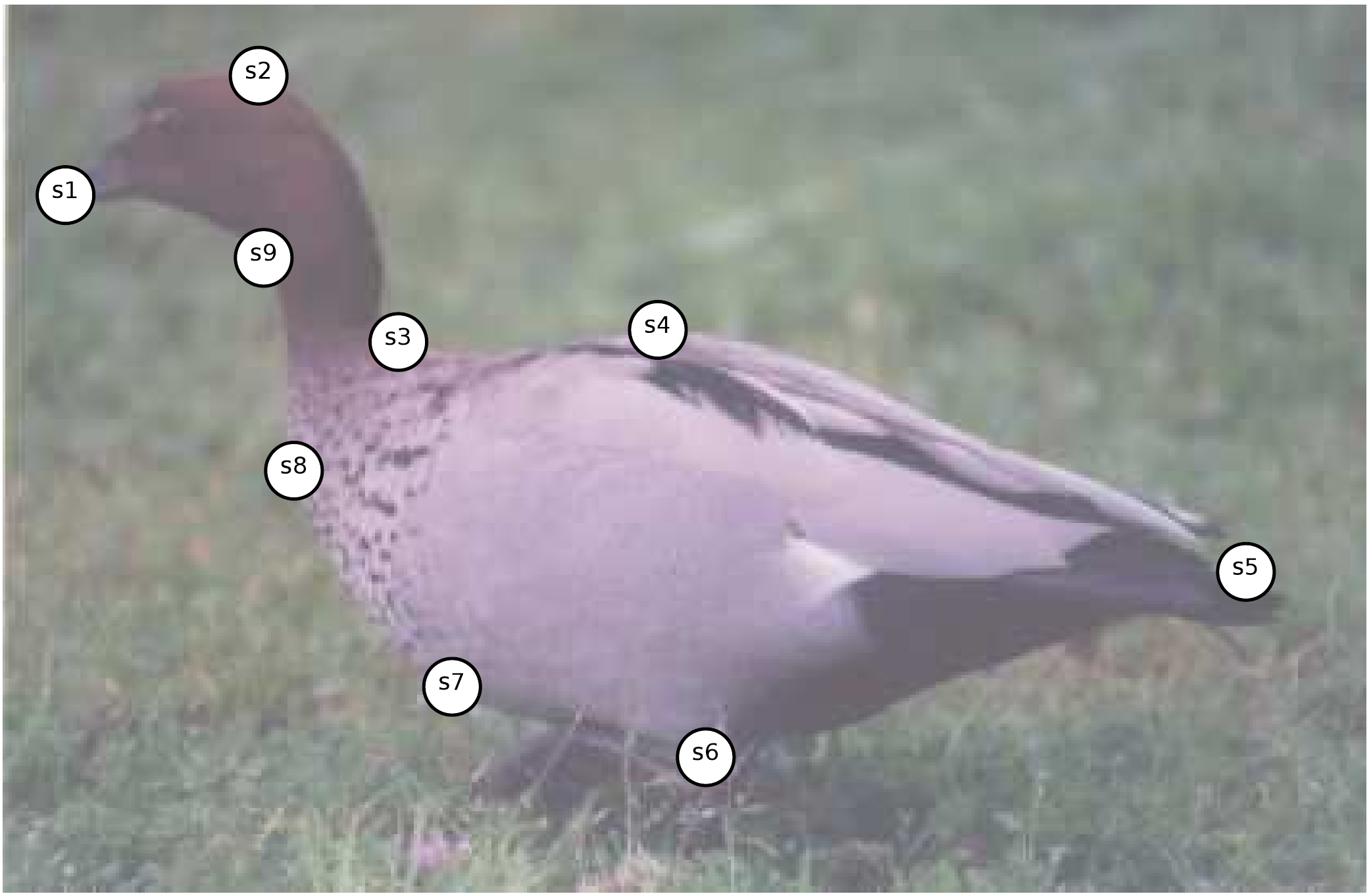}
\includegraphics[height=3.5cm]{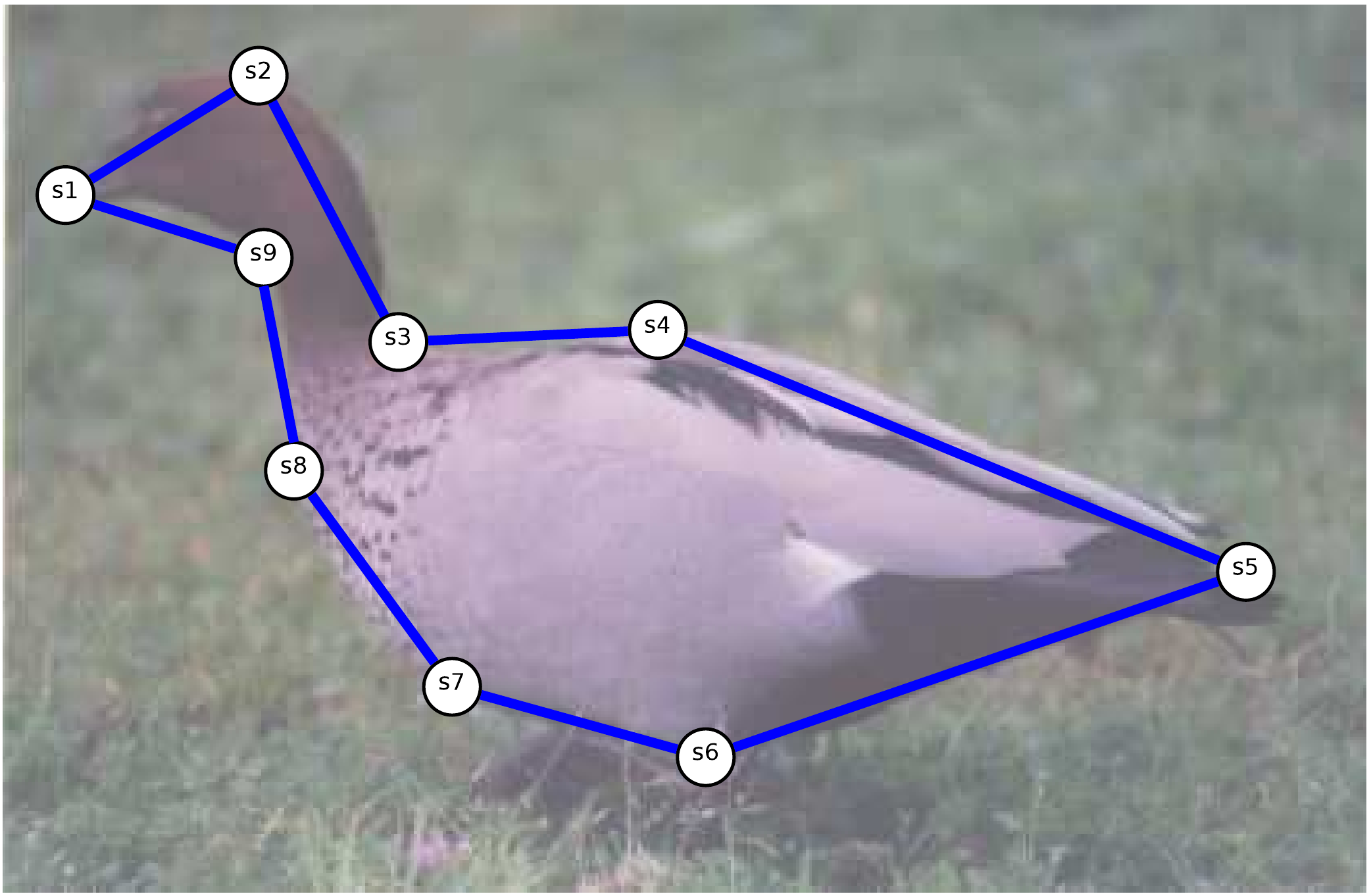}
\includegraphics[height=3.5cm]{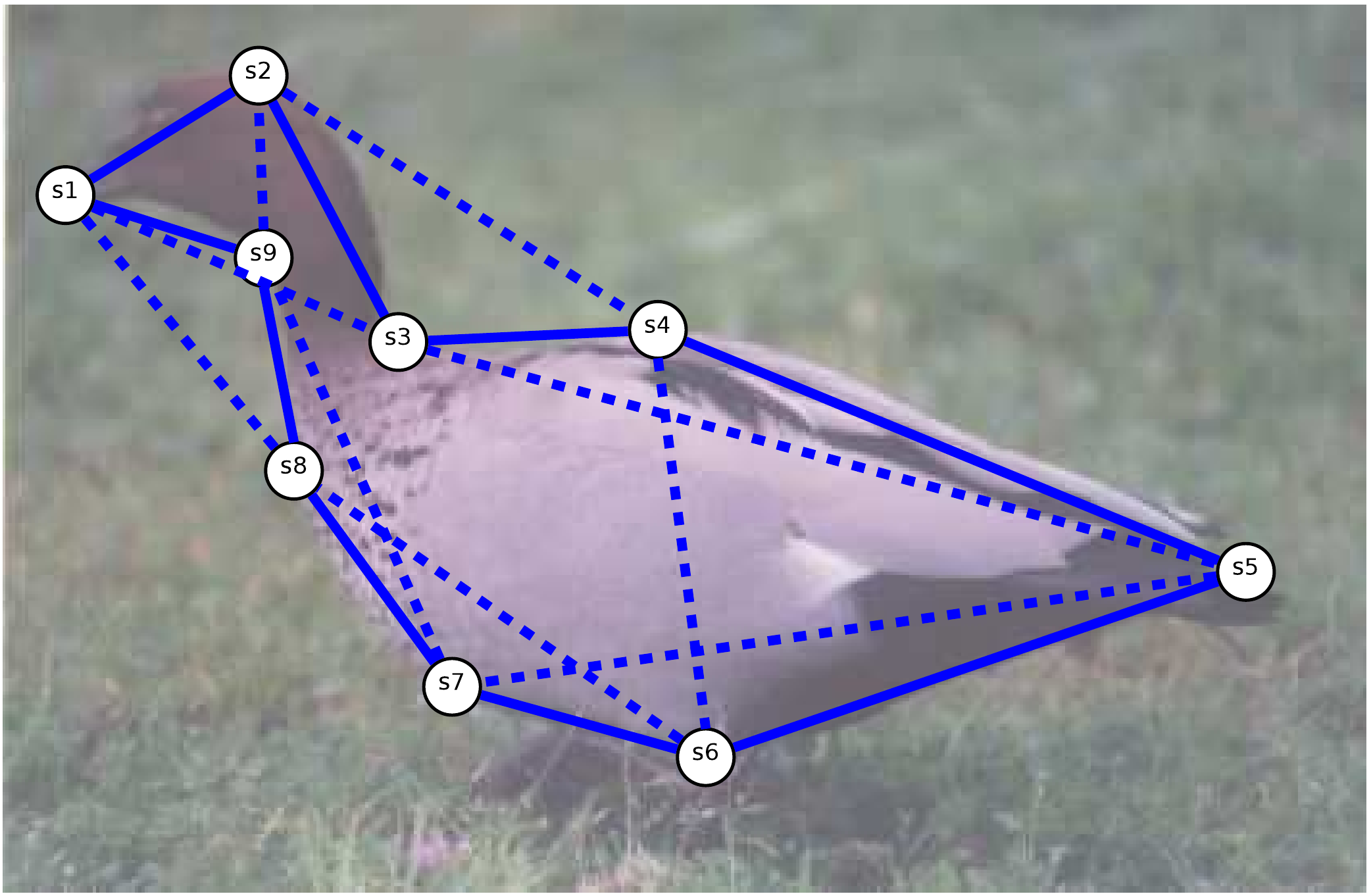}
\caption{Left: the (ordered) set of points in our template shape ($\mathcal S$). Centre: connections between immediate neighbours. Right: connections between neighbour's neighbours (our graphical model).}
\label{fig:ducks}
\end{figure*}

Although the model of \cite{McACaeBar08} solves isometric matching problems optimally, it provides no guarantees for \emph{near}-isometric problems, as it only considers those compatibilities which form cliques in our graphical model. However, we are often only interested in the boundary of the object: if we look at the instance of the model depicted in figure \ref{fig:ducks}, it seems to capture exactly the important dependencies; adding additional dependencies between distant points (such as the duck's tail and head) would be unlikely to contribute to this model.

With this in mind, we introduce three new features (for brevity we use the shorthand $y_i = y(s_i)$):
\begin{description}
\item[$\Phi_1(s_1, s_2, y_1, y_2) = (d_1(s_1, s_2) - d_1(y_1, y_2))^2$,]\ \\where $d_1(a,b)$ is the Euclidean distance between $a$ and $b$, scaled according to the width of the target scene.
\item[$\Phi_2(s_1, s_2, s_3, y_1, y_2, y_3) = (d_2(s_1, s_2, s_3) - d_2(y_1, y_2, y_3))^2$,]\ \\where $d_2(a,b,c)$ is the Euclidean distance between $a$ and $b$ scaled by the average of the distances between $a$, $b$, and $c$.
\item[$\Phi_3(s_1, s_2, s_3, y_1, y_2, y_3) = (\angle(s_1, s_2, s_3) - \angle(y_1, y_2, y_3))^2$,]\ \\where $\angle(a,b,c)$ is the angle between $a$ and $c$, w.r.t. $b$.\footnote{Using features of such different scales can be an issue for regularisation -- in practice we adjusted these features to have roughly the same scale. For full details, our implementation is available at (\emph{not included for blind review}).}
\end{description}

We also include the unary features $\Phi_0(s_1, y_1) = (\phi(s_1) - \phi(y_1))^2$ (i.e.~the pointwise squared difference between $\phi(s_1)$ and $\phi(y_1)$). $\Phi_1$ is exactly the feature used in \cite{McACaeBar08}, and is invariant to isometric transformations (rotation, reflection, and translation); $\Phi_2$ and $\Phi_3$ capture triangle similarity, and are thus also invariant to scale. In the context of (eq.~\ref{eq:negcost}), we have
\begin{multline}
\Phi(s_1, s_2, s_3, y_1, y_2, y_3) := \bigl[ \Phi_0(s_1, y_1),\\ \Phi_1(s_1, s_2, y_1, y_2) + \Phi_1(s_1, s_3, y_1, y_3),\\
\Phi_2(s_1, s_2, s_3, y_1, y_2, y_3) + \Phi_2(s_1, s_3, s_2, y_1, y_3, y_2),\\ \Phi_3(s_1, s_2, s_3, y_1, y_2, y_3) \bigr].
\end{multline}
This demands some explanation: only two pairwise dependencies ($\Phi_1$) are included in each clique -- this is done to ensure that each pairwise dependency is included exactly once, as the remaining dependency is captured by an adjacent clique. Furthermore, we have included \emph{two} scaled distances and \emph{one} angle ($\Phi_2$ and $\Phi_3$) -- although we could have included as many as \emph{three} scaled distances and \emph{three} angles, we have instead included exactly what is required to capture triangle similarity. Finally, we have enforced that features of the same type are given the same weight ($\Phi_1$ and $\Phi_2$), simply by adding the different instances of these features.

In practice, landmark detectors often identify several hundred points \cite{lowe99object,MikSch04}, which is clearly impractical for an $O(|\mathcal S| |\mathcal U|^3)$ method ($|\mathcal U|$ is the number of landmarks in the target scene). To address this, we adopt a two stage learning approach: in the first stage, we learn only unary compatibilities, exactly as is done in \cite{CaeCheLeSmo07}. During the second stage of learning, we collapse the first-order feature vector into a single term, namely
\begin{equation}
\Phi_0'(s_1, y_1) = \langle \theta_0,\Phi_0(s_1, y_1)\rangle
\end{equation}
($\theta_0$ is the weight vector learned during the first stage). We now perform learning for the third-order model, \emph{but consider only the $p$ `most likely' matches for each node}, where the likelihood is simply determined using $\Phi_0'(s_1, y_1)$. This reduces the memory and runtime requirements to $O(|\mathcal S| p^3)$. A consequence of using this approach is that we must now tune \emph{two} regularisation constants; this is not an issue in practice, as learning can be performed quickly using this approach.\footnote{In fact, even in those cases where a single stage approach was tractable (such as the experiment in section \ref{sec:house}), we found that the two stage approach worked better. Typically, we required much less regularity during the second stage, possibly because the higher order features are heterogeneous.}

\section{Experiments}
\label{sec:experiments}

\subsection{House Data}
\label{sec:house}

In our first experiment, we compare our method to those of \cite{McACaeBar08} and \cite{CaeCheLeSmo07}. Both papers report the performance of their methods on the CMU `house' sequence -- a sequence of 111 frames of a toy house, with 30 landmarks identified in each frame.\footnote{\scriptsize\texttt{http://vasc.ri.cmu.edu/idb/html/motion/house/index.html}} As in \cite{CaeCheLeSmo07}, we compute the Shape Context features for each of the 30 points \cite{BelMal00}.

\begin{figure*}
\begin{center}
\parbox{0.22\textwidth}
{
 \includegraphics[width=0.22\textwidth]{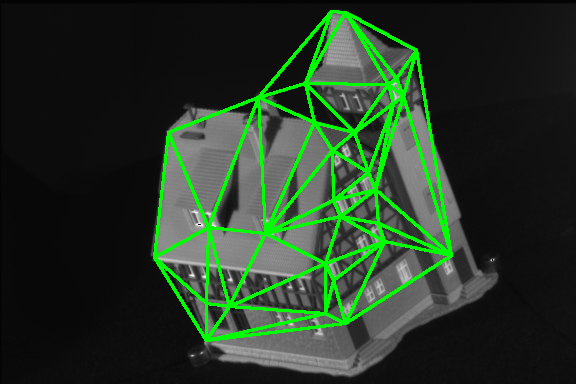}
 \includegraphics[width=0.22\textwidth]{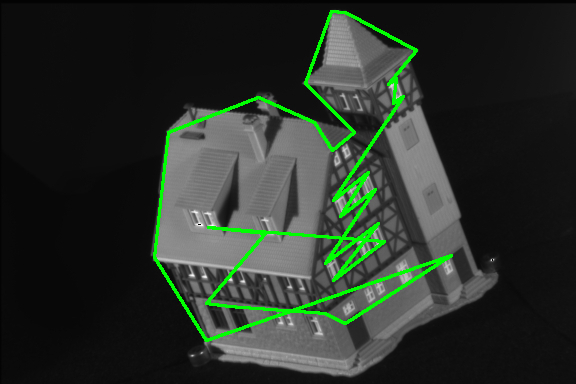}
 \includegraphics[width=0.22\textwidth]{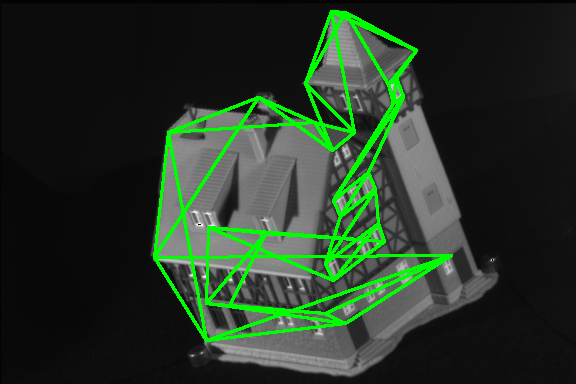}
}
\quad
\parbox{0.44\textwidth}
{
 \includegraphics[width=0.44\textwidth]{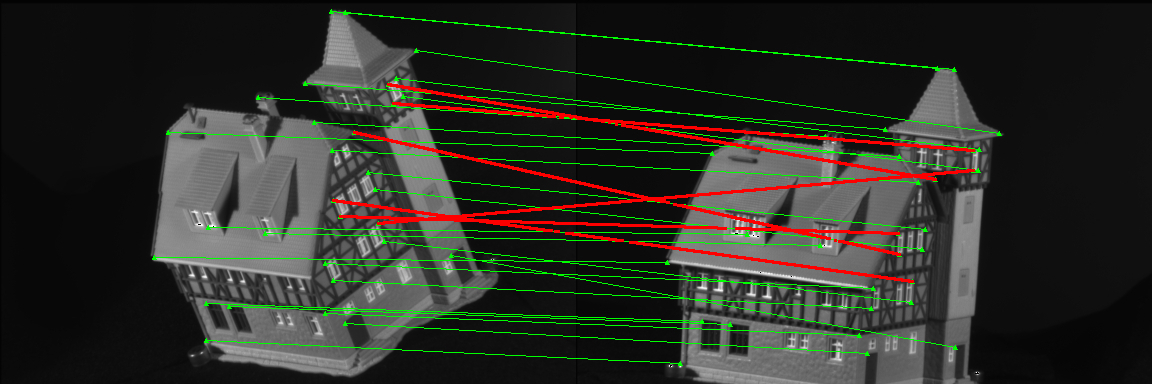}
 \includegraphics[width=0.44\textwidth]{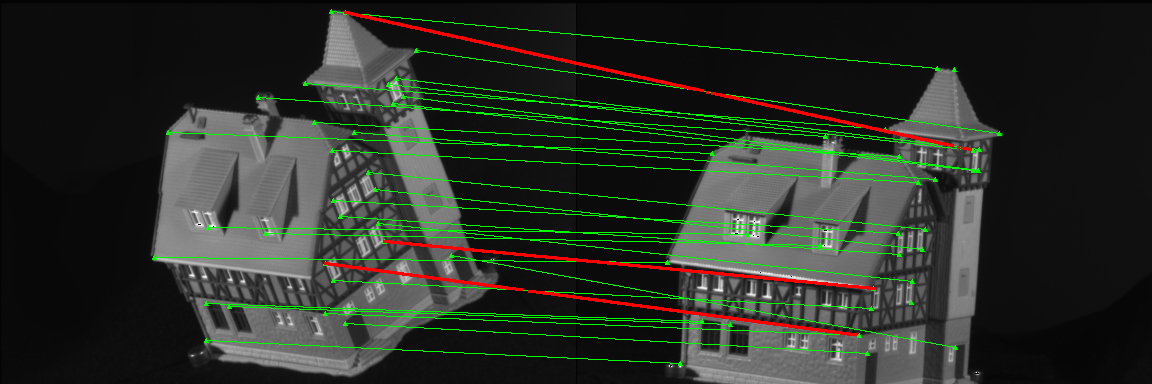}
 \includegraphics[width=0.44\textwidth]{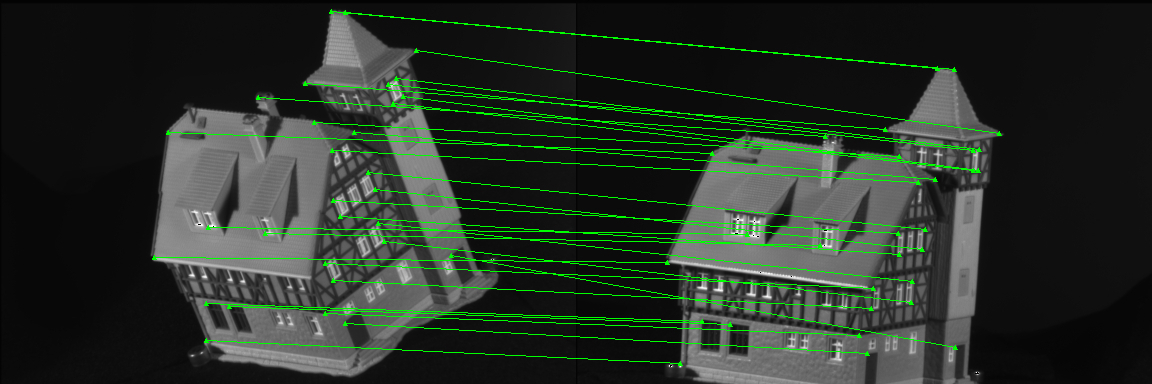}
}
\end{center}
\caption{Left: The adjacency structure of the graph (top); the boundary of our `shape' (centre); the topology of our graphical model (bottom). Right: Example matches using linear assignment (top, 6/30 mismatches), quadratic assignment (centre, 3/30 mismatches), and the proposed model (bottom, no mismatches). The images shown are the $3^{rd}$ and $93^{rd}$ frames in our sequence. Correct matches are shown in green, incorrect matches in red. All matches are reported \emph{after} learning.}
\label{housepics}
\end{figure*}

In addition to the unary model of \cite{CaeCheLeSmo07}, a model based on \emph{quadratic} assignment is also presented, in which pairwise features are determined using the adjacency structure of the graphs. Specifically, if a pair of points $(p_1, p_2)$ in the template scene is to be matched to $(q_1, q_2)$ in the target, there is a feature which is $1$ if there is an edge between $p_1$ and $p_2$ in the template, \emph{and} an edge between $q_1$ and $q_2$ in the target (and $0$ otherwise). We also use such a feature for this experiment, however our model only considers matchings for which $(p_1, p_2)$ forms an edge in our graphical model (see figure \ref{housepics}, bottom left). The adjacency structure of the graphs is determined using the Delaunay triangulation, (figure \ref{housepics}, top left).

\begin{figure}
\begin{center}
\parbox{\columnwidth}
{
\includegraphics[angle=-90,width=\columnwidth]{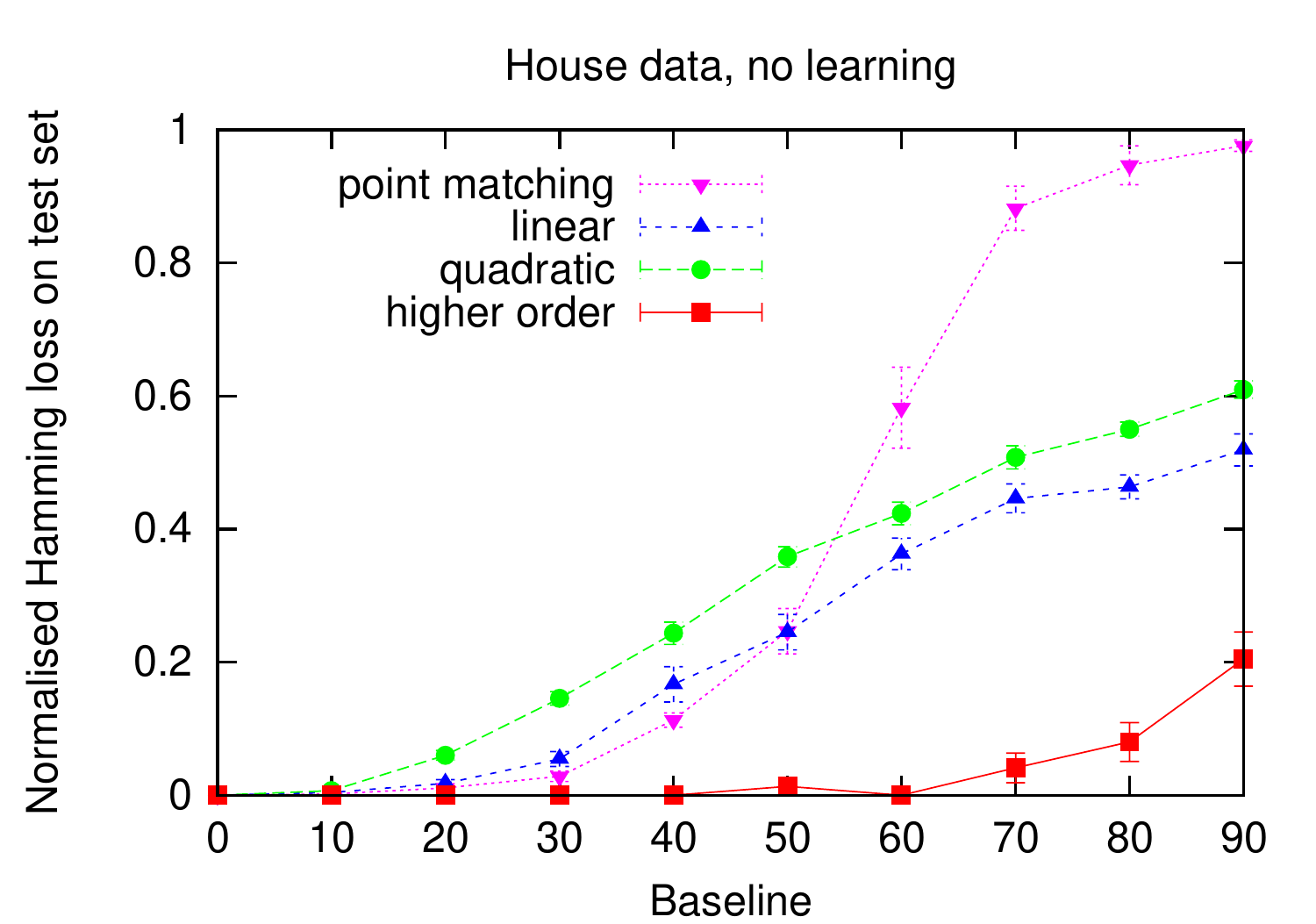}
\includegraphics[angle=-90,width=\columnwidth]{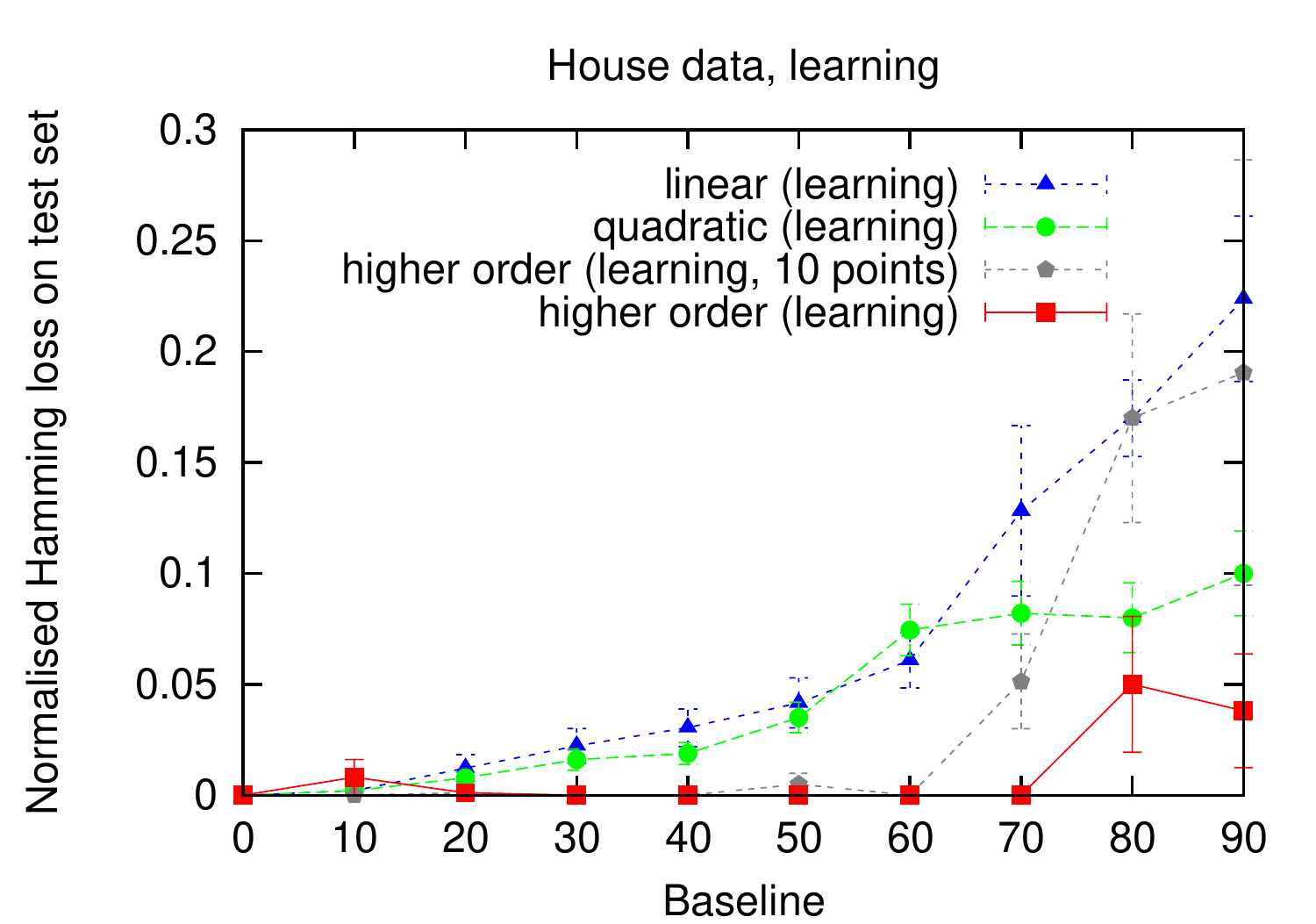}
}
\end{center}
\caption{Comparison of our technique against that of \cite{McACaeBar08} (`point matching'), and \cite{CaeCheLeSmo07} (`linear', `quadratic'). The performance before learning is shown on the top, the performance after learning is shown on the bottom. Our method exhibits the best performance both before and after learning (\emph{note the different scales of the two plots}). Error bars indicate standard error.}
\label{houseplot}
\end{figure}

As in \cite{McACaeBar08}, we compare pairs of images with a fixed baseline (separation between frames). For our loss function, $\Delta(\hat{y}, y^i)$, we used the normalised Hamming loss, i.e.~the proportion of mismatches. Figure \ref{houseplot} shows our performance on this dataset, as the baseline increases. On top we show the performance without learning, for which our model exhibits the best performance by a substantial margin.\footnote{Interestingly, the quadratic method of \cite{CaeCheLeSmo07} performs \emph{worse} than their unary method; this is likely because the \emph{relative} scale of the unary and quadratic features is badly tuned before learning, and is indeed similar to what the authors report. Furthermore, the results we present for the method of \cite{CaeCheLeSmo07} after learning are much \emph{better} than what the authors report -- in that paper, the unary features are scaled using a pointwise exponent ($-\exp(-|\phi_a-\phi_b|^2)$), whereas we found that scaling the features linearly ($|\phi_a - \phi_b|^2$) worked better.} Our method is also the best performing after learning (figure \ref{houseplot} (bottom))-- in fact, we achieve almost zero error for all but the largest baselines (at which point our model assumptions become increasingly violated, and we have less training data). 

\begin{figure}
\begin{center}
\includegraphics[angle=-90,width=\columnwidth]{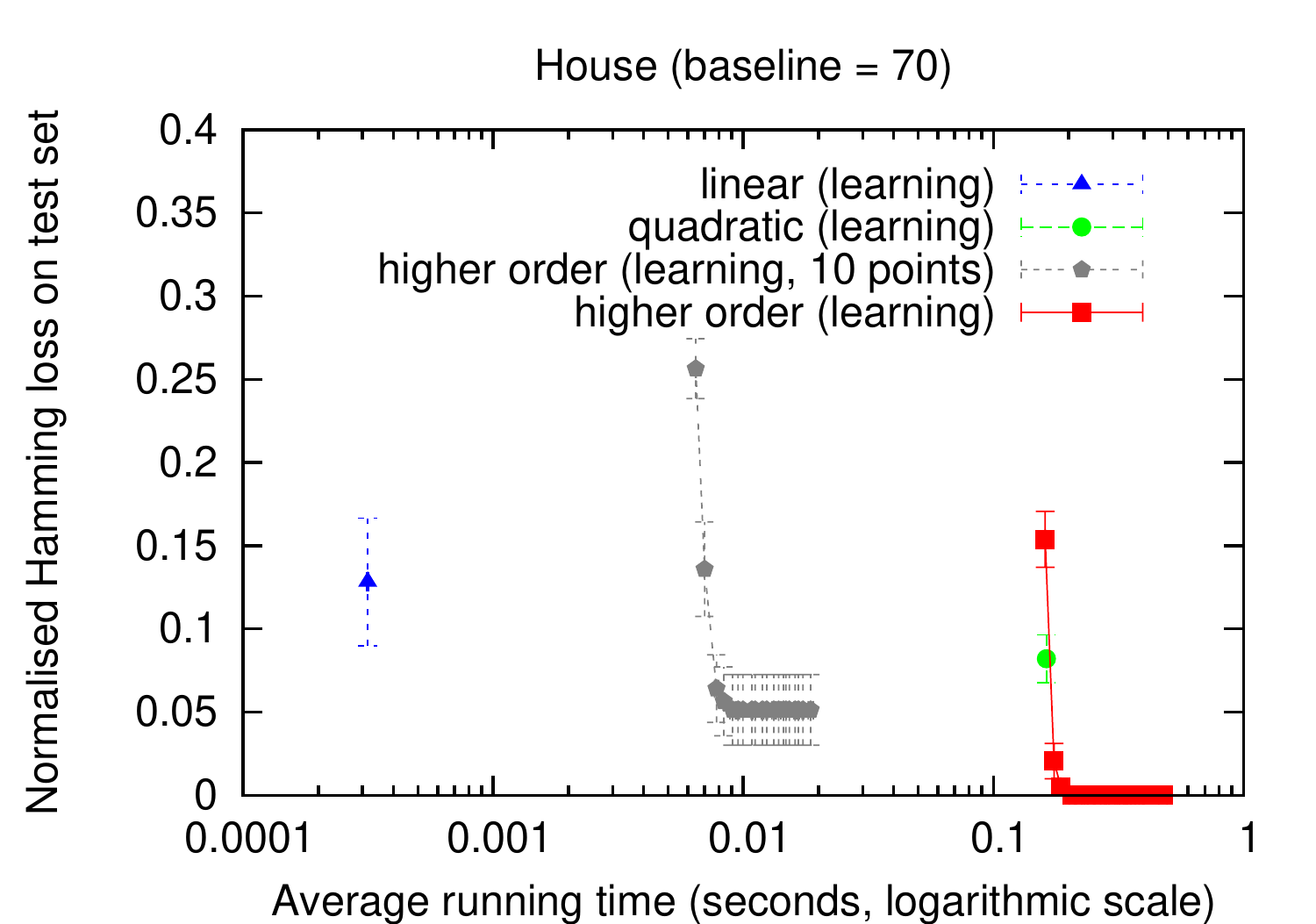}
\end{center}
\caption{The running time and performance of our method, compared to those of \cite{CaeCheLeSmo07} (note that the method of \cite{McACaeBar08} has running time identical to our method). Our method is run from 1 to 20 iterations of belief propagation, although the method appears to converge in fewer than 5 iterations.}
\label{housetime}
\end{figure}

In figure \ref{housetime}, we see that the running time of our method is similar to the quadratic assignment method of \cite{CaeCheLeSmo07}. To improve the running time, we also show our results with $p = 10$, i.e.~for each point in the template scene, we only consider the 10 `most likely' matches, using the weights from the first stage of learning. This reduces the running time by more than an order of magnitude, bringing it closer to that of linear assignment; even this model achieves approximately zero error up to a baseline of 60.

\begin{figure}
\begin{center}
\vspace{-10mm}
\parbox{\columnwidth}
{
\includegraphics[angle=-90,width=\columnwidth]{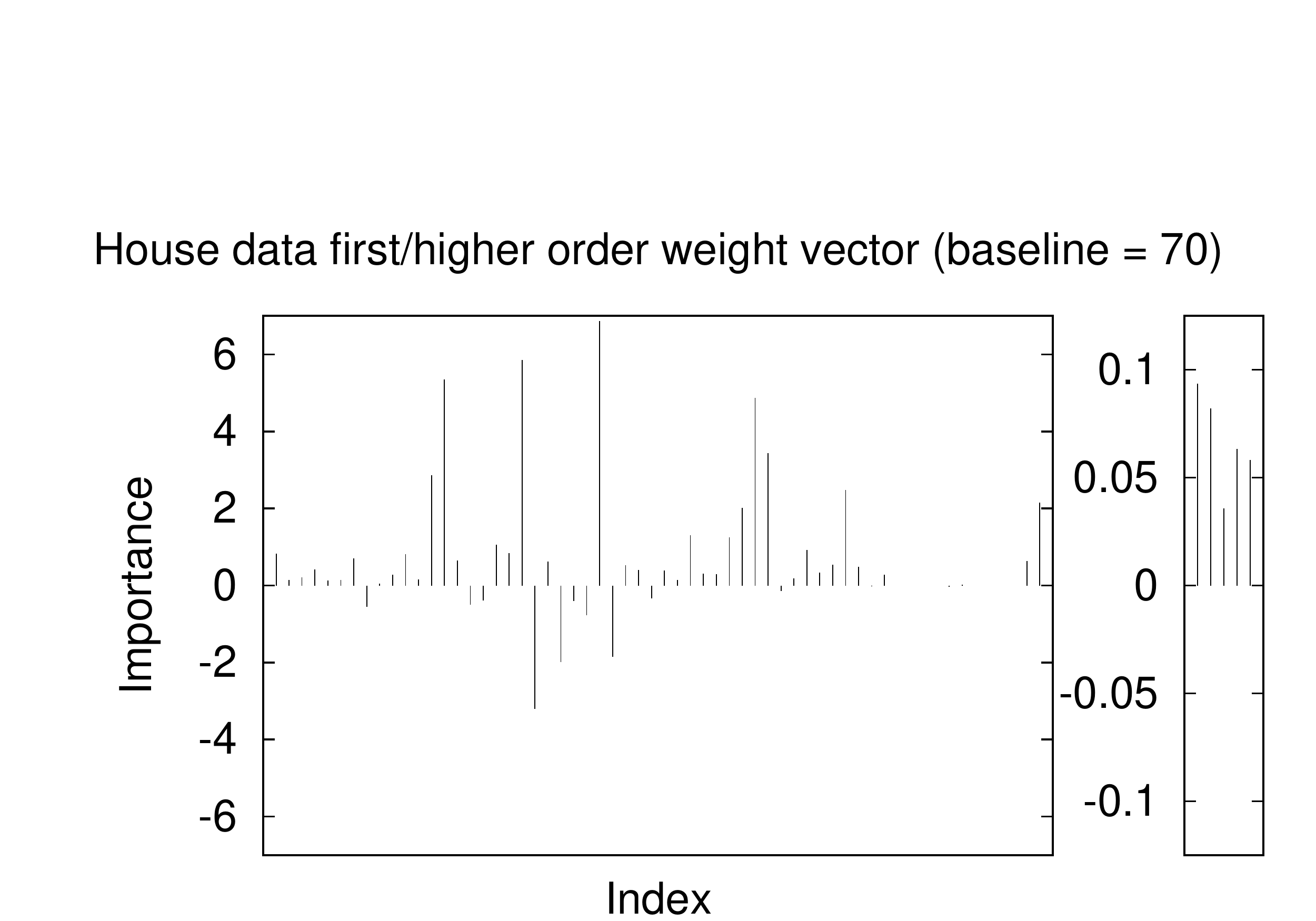}
\includegraphics[angle=-90,width=\columnwidth]{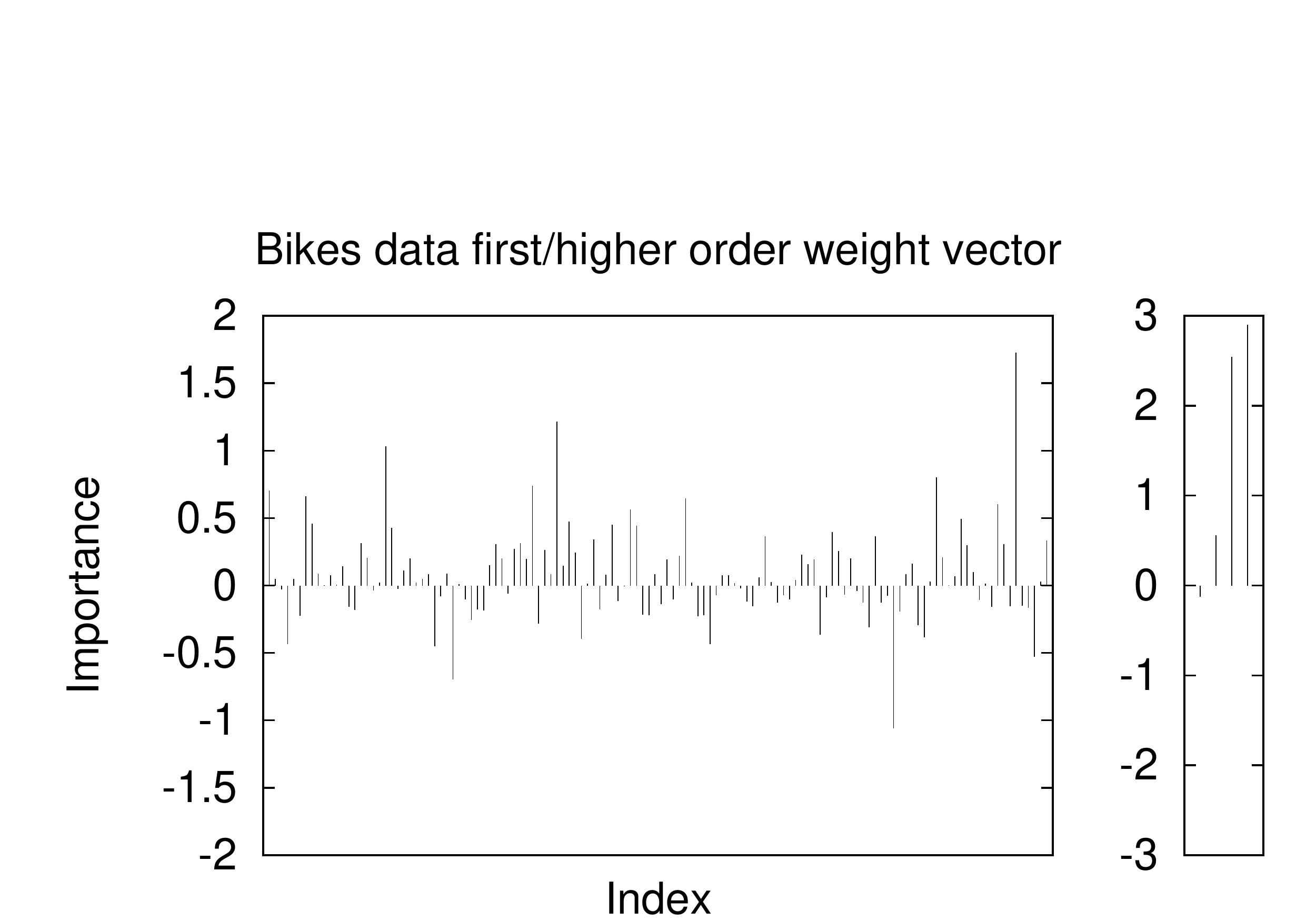}
}
\end{center}
\caption{Top: The weight vector of our method after learning, for the `house' data. The first 60 weights are for the Shape Context features from the first stage of of learning; the final 5 weights are for the second stage of learning. Bottom: The same plot, for the `bikes' data.}
\label{fig:weights}
\end{figure}

Finally, figure \ref{fig:weights} (top) shows the weight vector of our model, for a baseline of 70. The first 60 weights are for the Shape Context features (determined during the first stage of learning), and the final 5 show the weights from our second stage of learning (the weights correspond to the first-order features, distances, adjacencies, scaled distances, and angles, respectively -- see section \ref{sec:ourmodel}). We can provide some explanation of the learned weights: the Shape Context features are separated into 5 radial, and 12 angular bins -- the fact that there is a peak for the $14^{th}$, $26^{th}$, and $38^{th}$ features indicates that a particular angular bin is more important than the others; the fact that the final 12 features have low weight indicates that the most distant radial bin has little importance (etc.). It is much more difficult to reason about the second stage of learning, as the features have different scales, and cannot be compared directly -- however, it appears that all of the higher-order features are important to our model.

It is worth briefly mentioning that we also ran this experiment using our model, \emph{but including only the adjacency features, and ignoring all third-order features} -- i.e.~replicating exactly the experiment from \cite{CaeCheLeSmo07}, but including only the limited dependencies captured by our model. In this experiment, the model of \cite{CaeCheLeSmo07} performed better than ours; this indicates that the benefit of using an exact algorithm does not exceed the cost of capturing only limited dependencies. Indeed, this indicates that the third-order features are playing a very significant role in contributing to the performance of our model.

\subsection{Synthetic Data}

\begin{figure}
\begin{center}
\includegraphics[width=\columnwidth]{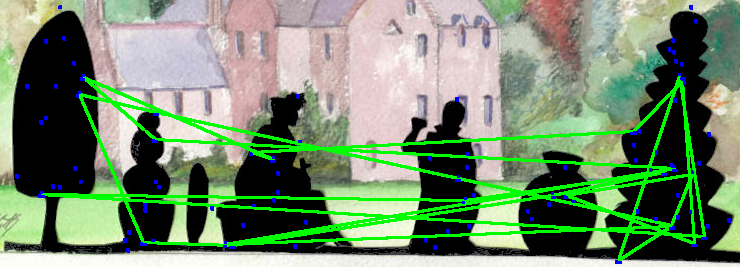}
\end{center}
\caption{The silhouette (from Steve Abbott's painting of Fingask Castle) on which our points are distributed. The `shape' of our model is shown in green, with outliers shown in blue.}
\label{fig:fingask}
\end{figure}

For this experiment, our `shape' consists of 25 points randomly distributed on the silhouette of the painting shown in figure \ref{fig:fingask} (note that this shape exhibits less structure than those in our other experiments, due to the random ordering of the points). In addition to the points on our shape, a number of outliers are randomly distributed on the silhouette. 10 training, testing, and validation images are then generated by randomly perturbing the $x$ and $y$-coordinates of these points by between $-\epsilon/2$ and $\epsilon/2$ pixels, where epsilon ranges between 0 and 20. This produces 45 \emph{pairs} of images for training, validation, and testing. This experiment is aimed at examining the robustness of our approach to noise and outliers, as well as the effect of choosing different values for $p$.\footnote{Note that setting $p=1$ essentially recovers the linear method of \cite{CaeCheLeSmo07}.}

\begin{figure*}
\begin{center}
\parbox{\columnwidth}{\includegraphics[angle=-90,width=\columnwidth]{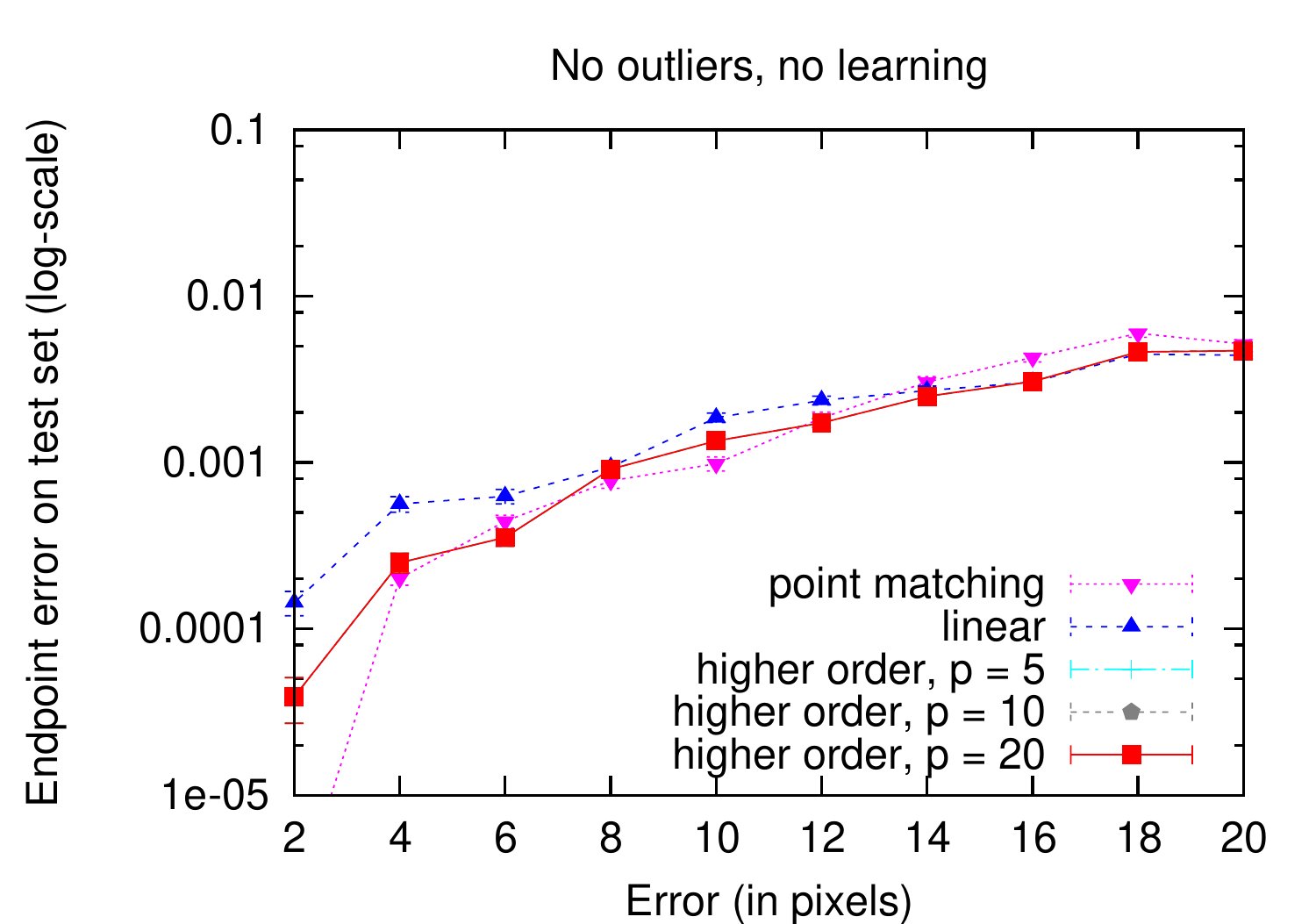}}
\parbox{\columnwidth}{\includegraphics[angle=-90,width=\columnwidth]{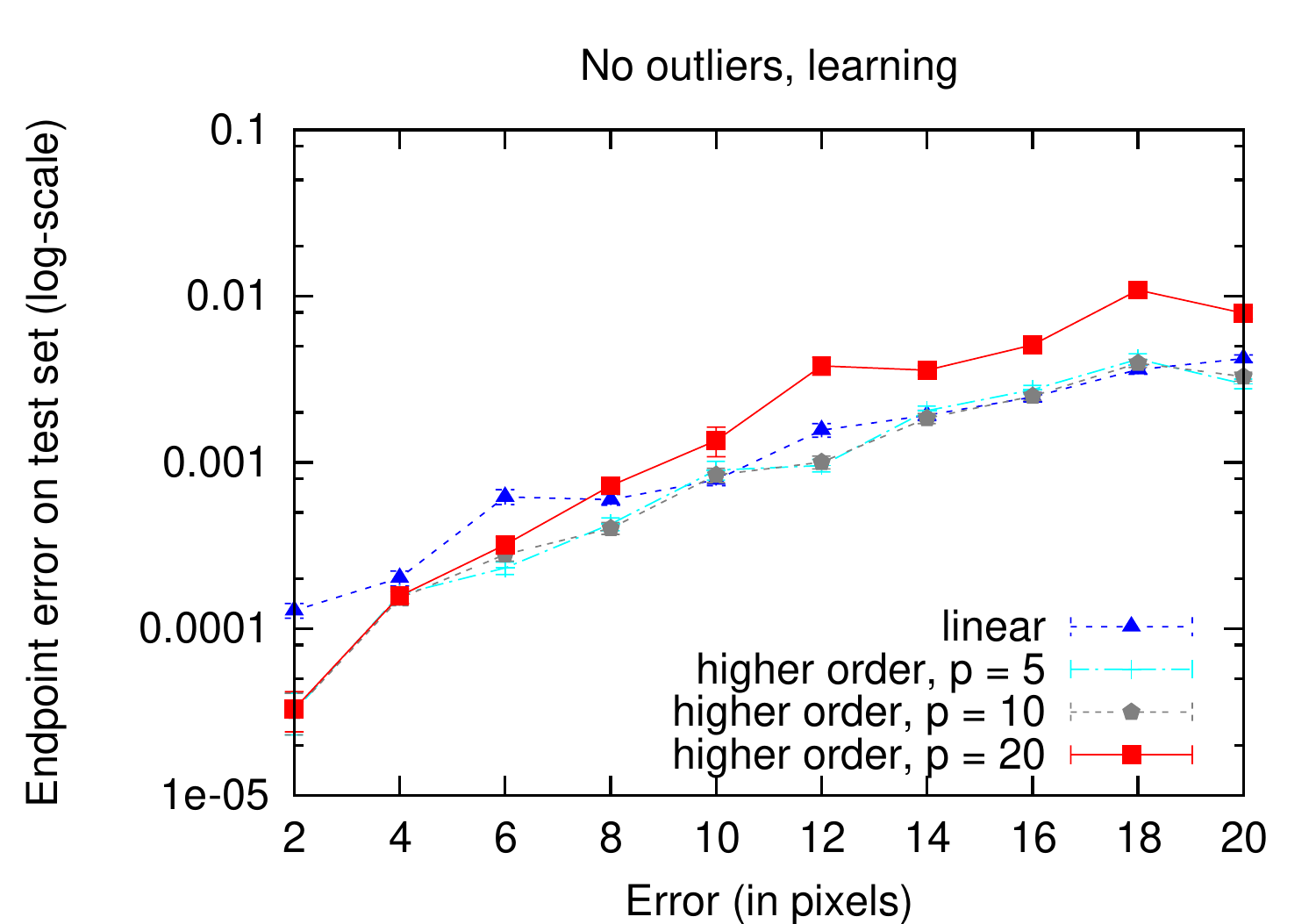}}
\parbox{\columnwidth}{\includegraphics[angle=-90,width=\columnwidth]{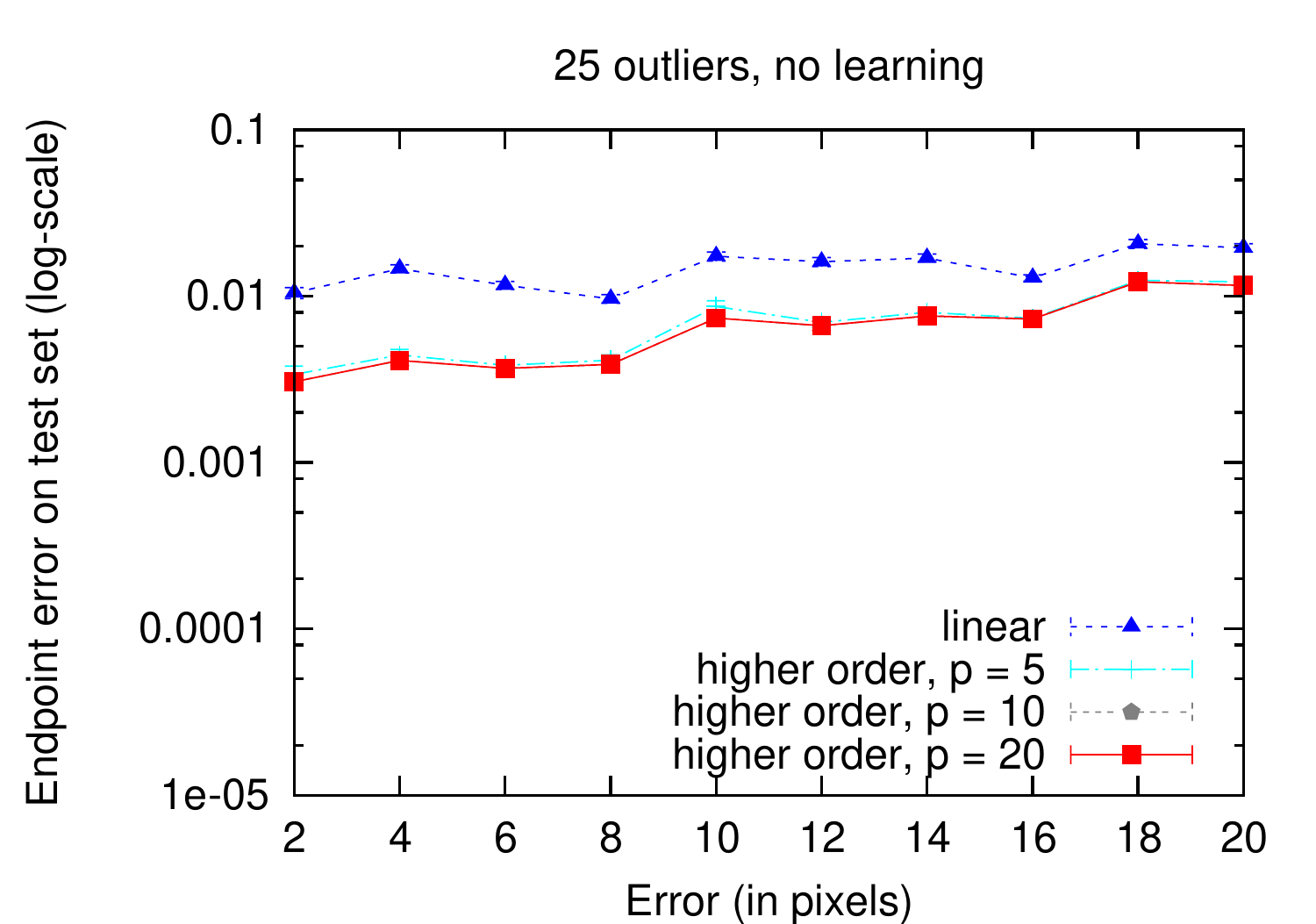}}
\parbox{\columnwidth}{\includegraphics[angle=-90,width=\columnwidth]{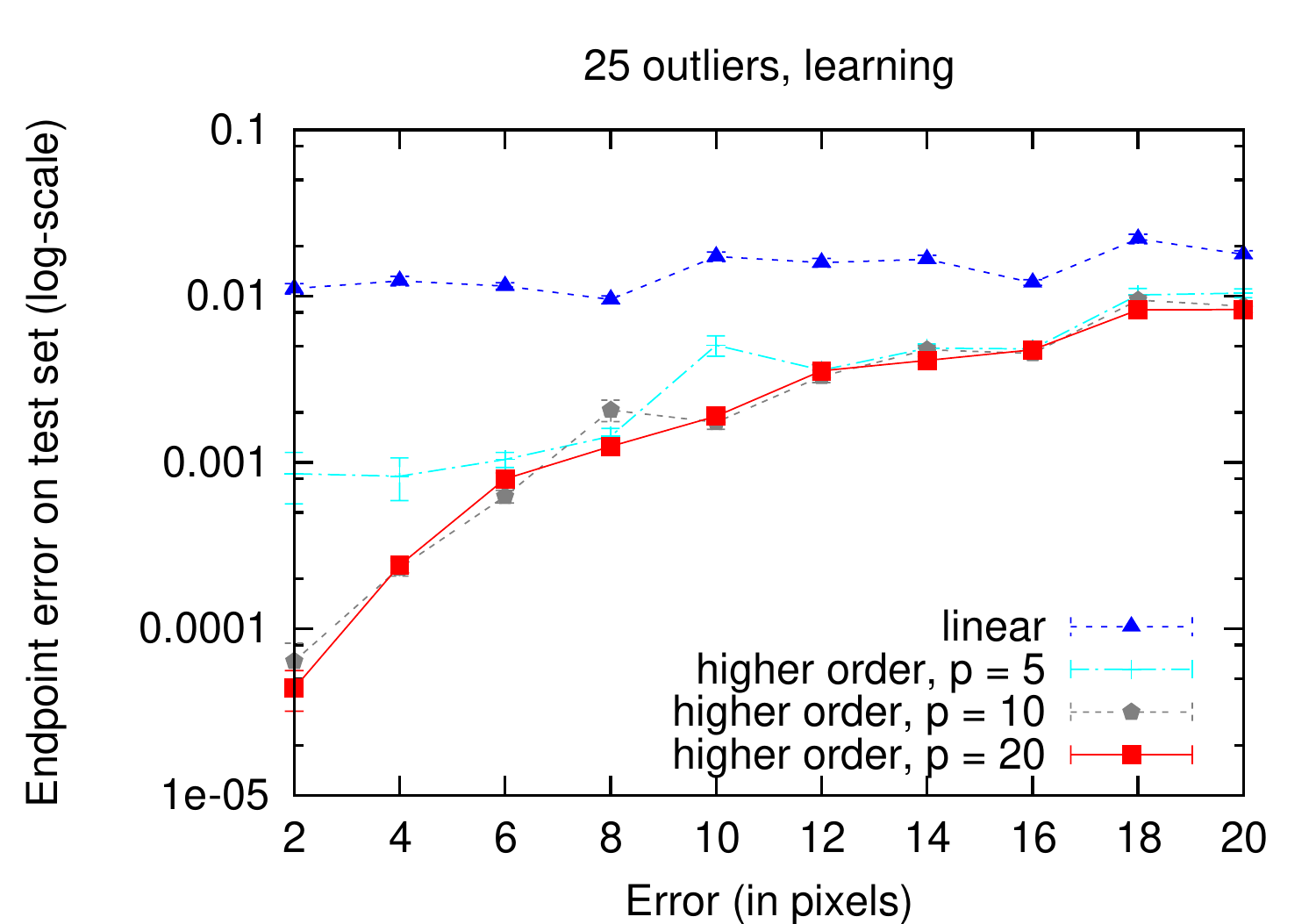}}
\parbox{\columnwidth}{\includegraphics[angle=-90,width=\columnwidth]{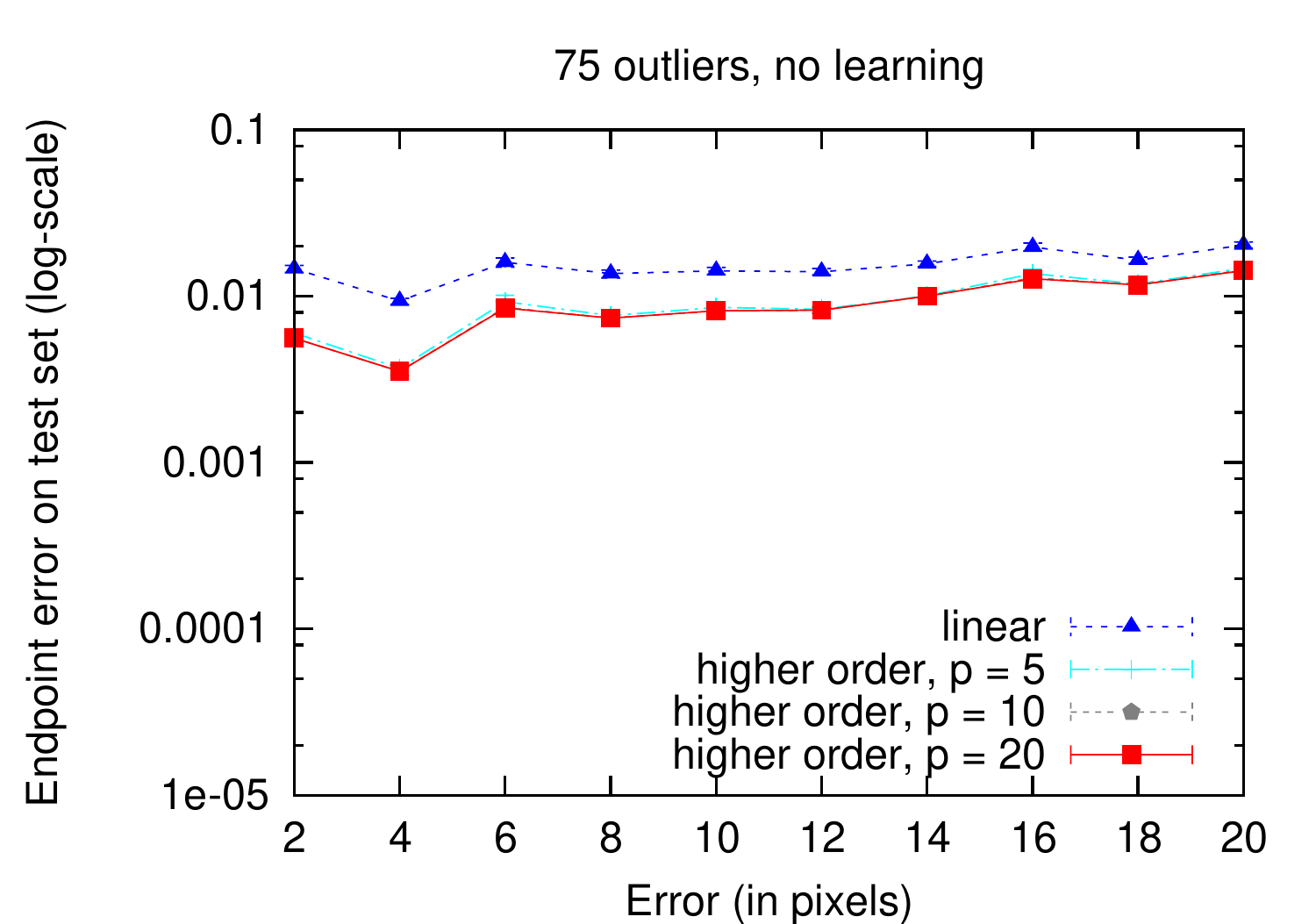}}
\parbox{\columnwidth}{\includegraphics[angle=-90,width=\columnwidth]{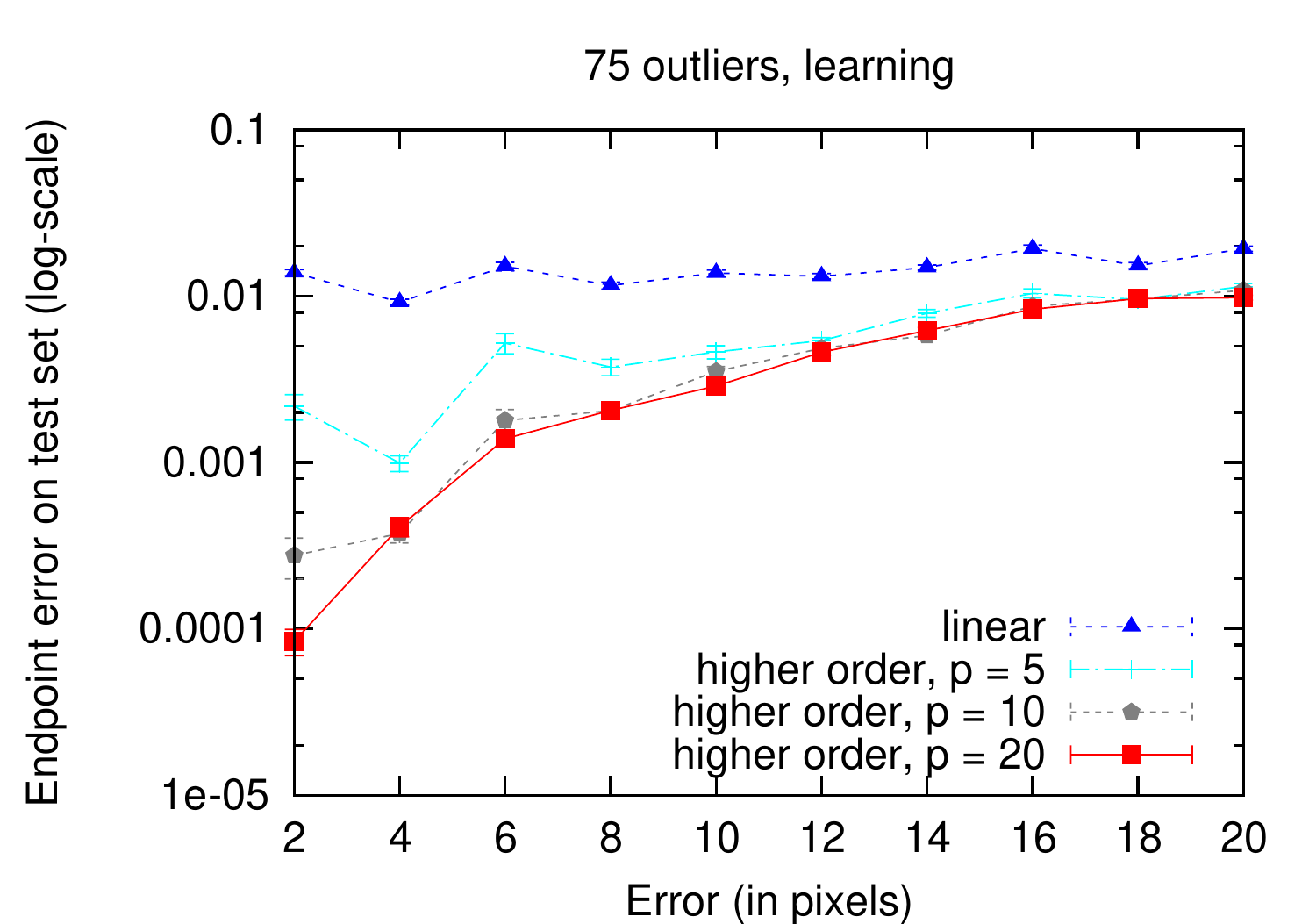}}
\end{center}
\caption{Comparison of our technique against that of \cite{McACaeBar08} (`point matching'), and \cite{CaeCheLeSmo07} (`linear'). Results are shown from errors ($\epsilon$) ranging from 2 to 20 pixels, for 0 (top), 25 (middle), and 75 (bottom) outliers. Results before learning are shown on the left, results after learning are shown on the right. Note the log-scale of the $y$-axis. Error bars indicate standard error. In many plots, the performance is almost identical for different values of $p$.}
\label{synthplot}
\end{figure*}

The results of this experiment are shown in figure \ref{synthplot}. Note that the `point matching' method is only shown for zero outliers, as the method becomes intractable as $|\mathcal U|$ increases. The quadratic assignment method of \cite{CaeCheLeSmo07} is not shown for this experiment, as the adjacency information in the graph is not robust to random error, or the addition of outliers (it performed far worse than the techniques shown). Since we cannot hope to get exact matches, we use the endpoint error instead of the normalised Hamming loss, i.e.~we reward points which are close to the correct match.\footnote{Here the endpoint error is just the average Euclidean distance from the correct label, scaled according to the width of the image.} Figure \ref{synthplot} also examines the effect of choosing different values of $p$ (the number of points considered during the second stage of learning).

Given that our datapoints are generated randomly, we observe little improvement from learning when using first-order features. Although the higher-order model provides no benefit when there are no outliers, it is highly beneficial once outliers are introduced; we also observe a significant benefit from learning, which likely indicates that the \emph{relative} weights of the low and higher-order features are being adjusted. Finally, although we observe poor performance for $p=5$, we observe almost no difference when increasing $p$ from 10 to 20.

\subsection{Bikes Data}

\begin{figure*}
\begin{center}
\includegraphics[height=0.1\textwidth]{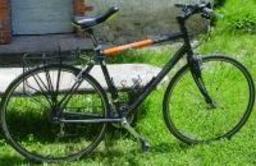}
\includegraphics[height=0.1\textwidth]{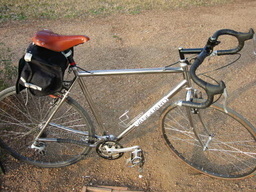}
\includegraphics[height=0.1\textwidth]{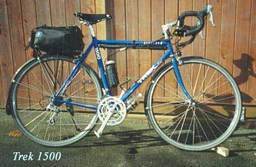}
\includegraphics[height=0.1\textwidth]{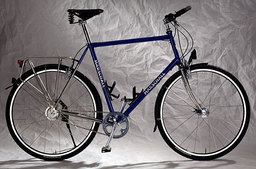}
\includegraphics[height=0.1\textwidth]{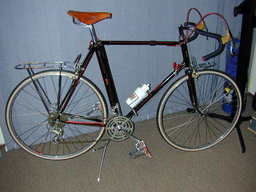}
\includegraphics[height=0.1\textwidth]{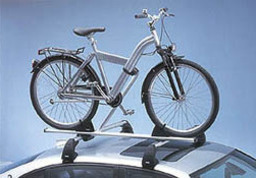}
\\
\vspace{1mm}
\includegraphics[width=\textwidth]{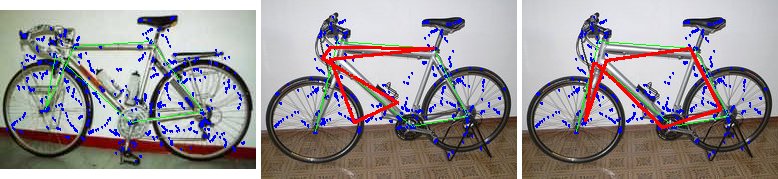}
\end{center}
\caption{Top: A selection of our training images. Bottom: An example match from our test set. Left: The template image (with the shape outlined in green, and landmark points marked in blue). Centre: The target image, and the match (in red) using unary features with the affine invariant/SIFT model of \cite{MikSch04} after learning (endpoint error = 0.34). Right: the match using our model after learning (endpoint error = 0.04).}
\label{fig:bikes}
\end{figure*}

For our final experiment, we used images of bicycles from the Caltech 256 Dataset \cite{caltech256}. Bicycles are reasonably rigid objects, meaning that matching based on their shape is logical. Although the images in this dataset are fairly well aligned, they are subject to reflections as well as some scaling and shear. For each image in the dataset, we detected landmarks automatically, and six points on the frame were hand-labelled (see figure \ref{fig:bikes}). Only shapes in which these interest points were not occluded were used, and we only included images that had a background; in total, we labelled 44 images. The first image was used as the `template', the other 43 were used as targets. Thus we are learning to match bicycles similar to the chosen template.

Initially, we used the SIFT landmarks and features as described in \cite{lowe99object}. Since this approach typically identifies several hundred landmarks, we set $p=20$ for this experiment (i.e.~we consider the 20 most likely points). Again we use the endpoint error for this experiment. Table \ref{tab:bikes} reveals that the performance of this method is quite poor, even with the higher-order model, and furthermore reveals no benefit from learning. This may be explained by the fact that although the SIFT features are invariant to scale and rotation, they are not invariant to reflection.

\begin{table*}
\caption{Performance on the `bikes' dataset. The endpoint error is reported, with standard errors in parentheses (note that the second-last column, `higher-order' uses the weights from the \emph{first} stage of learning, but not the second).}
\label{tablebikes}
\begin{center}
\begin{tabular}{|llcc|}
\hline
                   &             & SIFT \cite{lowe99object} & Affine invariant/SIFT \cite{MikSch04}\\
\hline
unary          & training:   & 0.335 (0.038)            & 0.321 (0.018)\\
               & validation: & 0.346 (0.027)            & 0.337 (0.015)\\
               & testing:    & 0.371 (0.011)            & 0.332 (0.024)\\
\hline
+ learning     & training:   & 0.277 (0.024)            & 0.286 (0.024)\\
               & validation: & 0.325 (0.020)            & 0.300 (0.020)\\
               & testing:    & 0.371 (0.011)            & 0.302 (0.016)\\
\hline
\hline
higher-order   & training:   & 0.233 (0.047)            & 0.205 (0.043)\\
               & validation: & 0.223 (0.025)            & 0.254 (0.035)\\
               & testing:    & 0.289 (0.045)            & 0.294 (0.034)\\
\hline
+ learning     & training:   & 0.254 (0.046)            & 0.211 (0.036)\\
               & validation: & 0.224 (0.025)            & 0.234 (0.035)\\
               & testing:    & 0.289 (0.045)            & \bf 0.233 (0.034)\\
\hline
\end{tabular}
\end{center}
\label{tab:bikes}
\end{table*}

In \cite{MikSch04}, the authors report that the SIFT features \emph{can} provide good matches in such cases, as long as landmarks are chosen which are locally invariant to affine transformations. They give a method for identifying affine-invariant feature points, whose SIFT features are then computed.\footnote{We used publicly available implementations of both methods.} We achieve much better performance using this method, and also observe a significant improvement after learning. Figure \ref{fig:bikes} shows an example match using both the unary and higher-order techniques.

Finally, figure \ref{fig:weights} (right) shows the weights learned for this model. Interestingly, the first-order term during the second stage of learning has almost zero weight. This must not be misinterpreted: during the second stage, the response of each of the 20 candidate points is so similar that the first-order features are simply unable to convey any new information -- yet they are still very useful in determining the 20 candidate points.

\section{Discussion and Future Work}

While our model seems well motivated when applied to the problem of `shape' matching (i.e.~when the shape has a clearly defined boundary), we are clearly making a tradeoff when applying our model to the more general problem of matching point-patterns. In such cases, we are at a disadvantage due to the fact that we capture only a fraction of the desired dependencies, but we are at an advantage in that our model is exact, and also that it is able to capture higher-order properties of the scene. Interestingly, we found that the exactness of our model alone does not make up for this limitation. This reveals the surprising result that the scale-invariant third-order features are able to capture a great deal of additional information that is not present at lower orders.

A hurdle faced by our approach is that of occlusions (either due to the landmark detector failing to identify part of the shape, or simply due to part of the shape being missing from the scene). Occlusions are of little concern to a first-order model, as an incorrect assignment to a single point has no effect on other assignments, whereas they may adversely effect our model, as the assignments are inextricably linked. In this paper, we have effectively dealt with the first issue (i.e.~that of the landmark detector failing to identify an important point), by using learning to select candidate landmarks. Dealing with occlusions explicitly is an important future addition to our model.

Another issue we encountered was that of feature scaling. For instance, suppose we express angles in degrees rather than radians; from the point of view of our model, this should make no difference -- we would just scale the corresponding weights by $\pi/180$; but from the point of view of the \emph{regulariser}, this is a \emph{very} significant change -- it is much more `expensive' to include a feature with a small scale (relative to other features) than it is to include a feature with a large scale. In theory, we would like to include many different features, and have the learning algorithm separate the good from the bad; in practice, this was not possible, as we were forced to address the relative scale of our features before we were able to do learning.\footnote{For full details, our implementation is available at (\emph{our implementation will be made available at the time of publication})} This appears to be a fundamental issue when applying learning to models with heterogeneous features, for which we are not aware of a principled solution.

In this paper we have used `off-the-shelf' landmark detectors, and only applied learning \emph{after} landmarks have been detected. Since we know the `type' of landmarks we want in advance (they are labelled in the template scene), it may be possible to apply learning to the landmark detector itself in order to further improve the performance of our model.

It would also be possible to allow for shapes which are rigid in some parts, but less so in others. For instance, although the handlebars, wheels, and pedals appear in similar locations on all bicycles, the seat and crossbar do not; we could allow for this discrepancy by learning a \emph{separate} weight vector for each clique.

\section{Conclusion}
\label{sec:conclusion}

We have presented a model for near-isometric shape matching which is robust to typical additional variations of the shape. This is achieved by performing structured learning in a graphical model that encodes features with several different types of invariances, so that we can directly learn a ``compound invariance'' instead of taking for granted the exclusive assumption of isometric invariance. Our experiments revealed that structured learning with a principled graphical model that encodes both the rigid shape as well as non-isometric variations gives substantial improvements, while still maintaining competitive performance in terms of running time.


\end{document}